# Machine Learning to study the impact of gender-based violence in the news media

**Hugo J. Bello[1,*], Nora Palomar[2], Elisa Gallego[3], Lourdes Jiménez Navascués[4] and Celia Lozano[5].**

[1]*Department of Applied Mathematics, Universidad de Valladolid, Soria, España.*
[2]*Servicio de Psiquiatría, Complejo Asistencial de Soria, Soria, España.*
[3]*Servicio de Medicina Preventiva, Hospital General Universitario de Ciudad Real, Ciudad Real, España.*
[4]*Universidad de Valladolid. Facultad de Ciencias de la Salud de Soria*
[5]*Department of Big Data and Business Intelligence, Sermes CRO, Madrid, Spain.*

**\*Corresponding author. Email:** hugojose.bello@uva.es **(H.J.B.)**

**ABSTRACT**
**While it remains a taboo topic, gender-based violence (GBV) undermines the health, dignity, security and autonomy of its victims. Many factors have been studied to generate or maintain this kind of violence, however, the influence of the media is still uncertain. Here, we use Machine Learning tools to extrapolate the effect of the news in GBV. By feeding neural networks with news, the topic information associated with each article can be recovered. Our findings show a relationship between GBV news and public awareness, the effect of mediatic GBV cases, and the intrinsic thematic relationship of GBV news. Because the used neural model can be easily adjusted, this also allows us to extend our approach to other media sources or topics.**

**INTRODUCTION**

Gender-based violence (GBV) is directly related at an individual based on her or his biological sex or gender identity [1,2]. It includes verbal, physical, sexual, and psychological abuse, whether occurring in public or private life. GBV is an issue faced by people all over the world, but women are disproportionately harmed [3,4]. Despite being so prevalent, gender-based violence is largely under-reported because of stigma and lack of access to resources and support systems. Only recently, the voices of victims, survivors and/or activists, have put the spotlight on the issue and raised public awareness by the media through movements such as #MeToo, #TimesUp, #Niunamenos, #NotOneMore, #BalanceTonPorc, and others.

News media are key to understanding how society and the general population react to topic [5]. This is prevalent with GBV, where news media are one of the main sources of information. Moreover, news can influence public perceptions and consequently, social policies [6,7]. Not only news media convey what happens, but also what could be done about it. They may report available help resources added to the piece of news. In the particular case of Spain, where more than a thousand women have been killed in the last two decades [8,9], many notorious cases have drowned the media attention as never before, for instance, "*the manada case*" [10]- a gang rape in 2016, lead to a series of protests all across the country-. This event (and its corresponding social, juridic and public conscience) has motivated us to study the reporting news in Spain. [11].

Here, we used Machine Learning, i.e. Natural Language Processing (NLP) [12], techniques to analyze millions of local news articles from the Spain media in order to extract the role of GBV comments on the media. NLP is a type of data analysis focused on transforming the free text (i.e. unstructured data) in documents and databases into normalized, structured data suitable for analysis. The potential contribution of Natural Language Processing to media articles has been recently explored [13–15]. To date, the efforts on NLP involve topic identification, natural language understanding, and natural language generation [16]. To obtain the main subjects of written media content, we use neural networks. Then, we search for topics related to gender violence and see how they are reported and how they are related. Our results are supported by three aspects: (i) the relationship between GBV news and public awareness, (ii) the effect of mediatic GBV cases, (iii) the intrinsic thematic relationship of GBV news.

## RESULTS

### Data Extraction and Topic Classification

This section describes the processing steps to extract the online news, transform unstructured text into structured data, and build NLP models for text classification, shown in the workflow of Figure 1a. Web scraping techniques were used to mine the data contained in Spanish media websites (in particular: elpais.com, elmundo.com, abc.com, lavanguardia.com, 20minutos.com, publico.es, and diario.es). Note that the first four are the most read online newspapers in Spain. The data covers 784 259 news from January 2005 to March 2020 (see Methods for further information). Each article contains a list of tags that allows its topic classification (such as *justice, economy, politics, international, technology, feminism*,...). Due to the lack of visibility of GBV, such a topic is not classified by the media. To bring light to this issue and, in addition, enhance the description tags, we extract for each news article its subject and how likely it is to be related to GBV. Therefore, we apply a stack protocol which consists of two consecutive neural network models for Natural Language Processing. (NN1 and NN2): firstly, by analyzing the content of each text, we extract the general subject classification, and secondly, we apply a binary gender-based violence classification (namely GBV probability which is a rate probability, where zero means lackness of GBV).

Firstly, we select a subsample of 300K news to train a topic classification model. This subgroup of news was already tagged by the different newspapers using common tags, which allows the neural network to learn from them, and classify each news with its corresponding keyword tags (see Methods NN1). Each news item could have several tags to describe the article. Fig. 1b shows the top 20 tags in the Spanish newspaper media. We point out: *politics, economy, Catalonia, elections* or*, jury*, which resemble the most relevant topics in society. Remarkably, none of these tags is directly related to gender.

Secondly, in order to shed light on the unspokeness of GBV, we categorize if the news covers gender-based violence or not. For sake of clarity, here we refer to GBV on a broad meaning which includes the tags feminices, domestic violence, female sexual abuse or/and sexual crimes against women. To train the model, we used 4K news which has at least one of its tags. The details of this model are explained in the Methods section (see Methods NN2). From such classification, for each new, we obtain a probability (a number between 1 and 0) that represents how much this new covers the subject of GBV.

In the following, we explore three aspects of the finding tags: (i) the relationship between GBV news and public awareness, (ii) the effect of mediatic GBV cases, (iii) the intrinsic thematic relationship of GBV news.

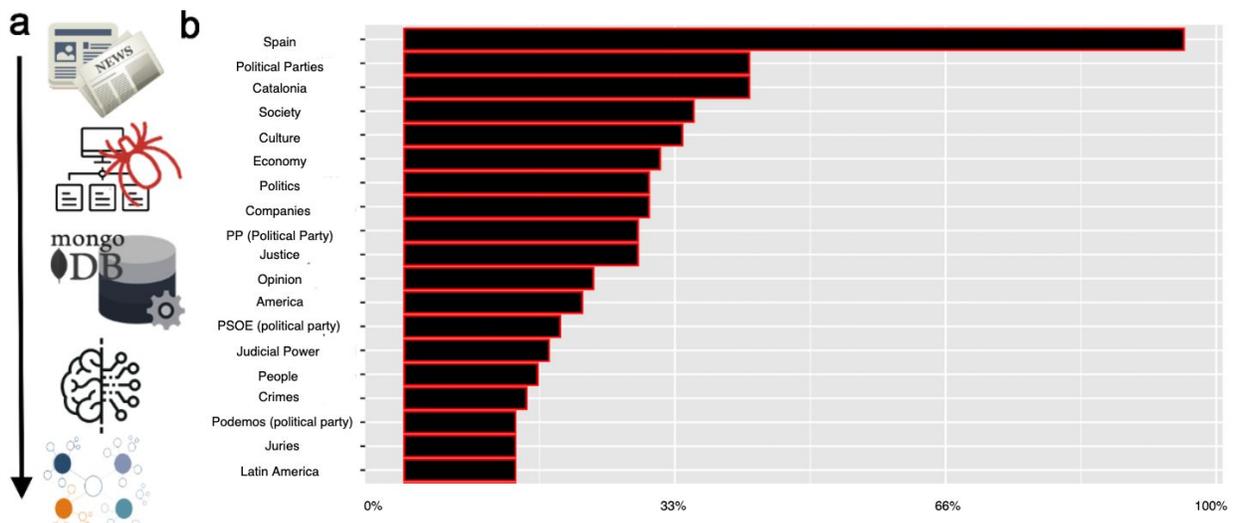

Figure 1. **Data Extraction and Topic Classification. (a)** Workflow of the news extraction process: (i) online news, (ii) web scraping, (iii) database store, (iv) data processing by neural network, (v) data analysis (e.g. Topological Data Analyse)  **(b)** Top 20 Topic Classification of news articles.

**Gender violence in news and public awareness**

To understand Gender-Based Violence (GBV) press coverage in more detail, we analyze the GBV probability over time, i.e. the temporal changes [17,18]. Considering the whole news, we measure the temporal monthly average of GBV probability (obtained with the model NN2 described in methods). The larger the probability, the greater its chances to talk about GBV on a certain month. The idea is straightforward: represent a time-series $X_t$ as a combination of trend $T_t$, seasonal $S_t$, and irregular components, or errors, $e_t$ using moving averages with the model:

$$X_t = T_t + S_t + e_t \qquad (1)$$

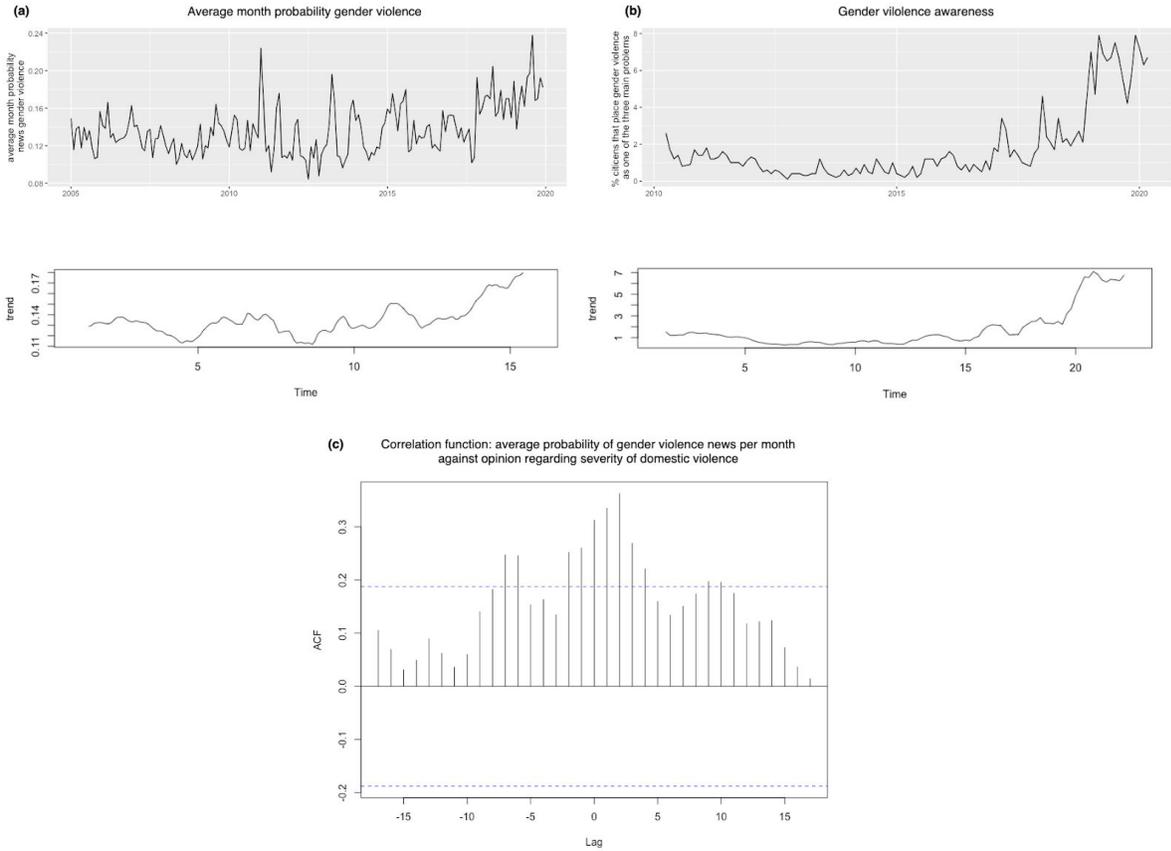

Figure 2. **GBV Time series.** (a) Monthly GBV probability and its corresponding trend. (b) Monthly percent of citizens that place gender violence as one of the three main problems (Centre for Sociological Research data) and its corresponding trend. (c) Cross-Correlation Function of the Temporal GBV and the Temporal feedback opinion.

Figure 2(a) shows that GBV probability is enhancing over time, in particular, by a factor of 1.8 in the last 3 years. Remarkably, 20% of the news in 2020 has GBV connotation. We want to compare such enhancement with public awareness concerning gender movements. For that, we take a look at the social perception of GVB using the monthly survey CIS ('Centre for Sociological Research' a Spanish public research institute)[19]. We apply the same seasonal-trend decomposition model (1), which gives us Figure 2(b). Comparing both sides of Figure 2 we can point out that monthly GBV probability in news and and gender violence awareness have the same upward trend.

Furthermore, quantitative information about the consistent trend behaviour is obtained from the time correlation between average GBV probability against the opinion survey about domestic violence as a function of the displacement of one relative to the other (see Fig. 2c). We find that the public opinion lags behind the media news for several months, which means that, after a relevant GBV event, people respond to it.

**The impact of mediatic gender violence cases**

Our results show that GBV probability monthly grouped has increased over the years. Similar trend is also shown in the daily grouped data. In order to get more insights, we focus on the daily GBV probability by decomposing the time series $Y_t$ with three main model components [20]: trend, seasonality, and holidays:

$$Y_t = T_t + S_t + H_t + e_t \qquad (2)$$

where $T_t$ is the trend, $S_t$ is the seasonal component, $H_t$ is the holiday component and $e_t$ is the error. To fit the data, we use Prophet model [21,22] (a Facebook's open software), in which $T_t$ can be either a rating growth model, or a piecewise linear model, $S_t$ is built using Fourier series and $H_t$ is set using the known holidays. Figure 3 shows the Prophet fit (green lines) of the GBV data (black symbol). Although it seems that the data fits well in the plot, certain days are out-of-the-range. So, this model can be used for anomaly detection as well: every value of the time series outside a 99% uncertainty interval of the fitting-model is considered an anomaly (red symbols in Fig.3).

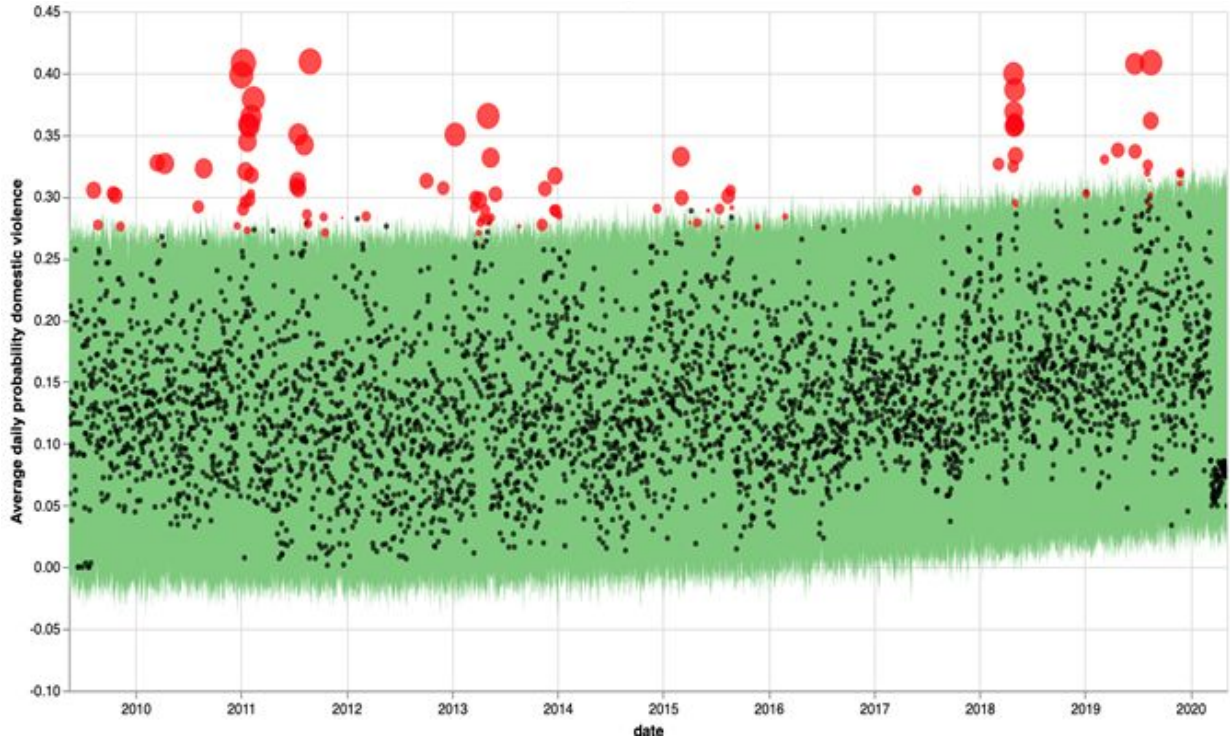

**Figure 3. Anomalies detection.** Prophet fit of the temporal GBV probability where solid symbols are the GBV data, green lines are the fitting curve, and the red symbols the anomalies points.

Table 1 shows the list of the most recent anomalies detected by the previous analysis (marked as red points in figure 3), the corresponding average GBV probabilities, and the event in the news. The match between the anomaly detection and a female violence case confirms that the Prophet algorithm applied to our data is a good method to detect GBV news. Note that several of the showing events correspond to the sentence Court of 'La Manada rape case' [10,23,24] where the gang rape of an 18-year-old woman on 7 July 2016 during the San Fermín celebrations in Pamplona, Spain.

| Date | GBV Average Probability | Event |
|---|---|---|
| 2019-08-17 | 0.4085262 | The news cover the murder of a female surgeon in a gender violence case. https://www.elmundo.es/madrid/2019/08/17/5d57c84a21efa0ab3b8b45cb.html |
| 2019-06-22 | 0.4075814 | In regards to the manada case, the Supreme Court of Spain upgraded convictions for sexual abuse to that of continuous sexual assault, and handed down 15-year prison terms. |
| 2018-05-02 | 0.3866444 | News covered the case of a gender violence based victim that was killed in Guadalix. *https://elpais.com/politica/2018/02/05/actualidad/1517833337_571770.html* |
| 2018-04-29 | 0.3687508 | The five members of the manada where condemned to nine years on the charges of sexual abuse. |
| 2018-04-28 | 0.3995969 | The five members of the manada where condemned to nine years on the charges of sexual abuse. |

**Table 1.** List of recent anomalies, the corresponding average probabilities and the event

In order to visualize the anomalies and the dates of high GBV probabilities in more detail, we use a heatmap plot. In Figure 4, we plot all the dates of the years 2018, 2019 and 2020 ( warmer color for the high average GBV probabilities). Surprisingly, the high probabilities of GBV not only correspond to the anomaly day, but appear over the course of several days. This result is reasonable because when an impacting GBV case occurs, this is talked about for several days. In addition, the heatmap shows how the years get hotter and hotter (in terms of probability of gender violence news), which confirms the upward trend that we saw in the time series (Figure 2).

To conclude that our anomalie model indeed captures the likelihood of gender violence news.

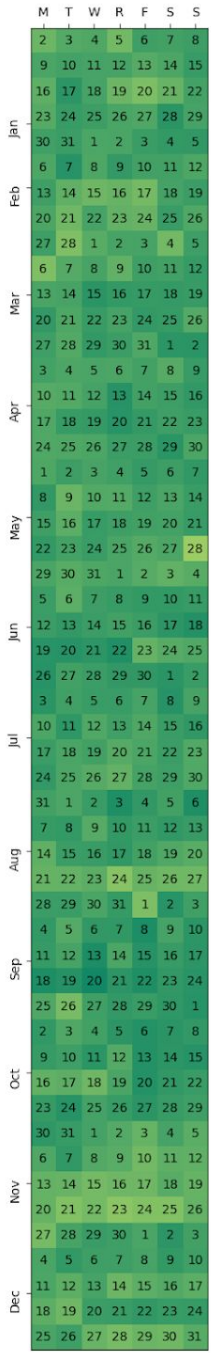
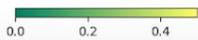
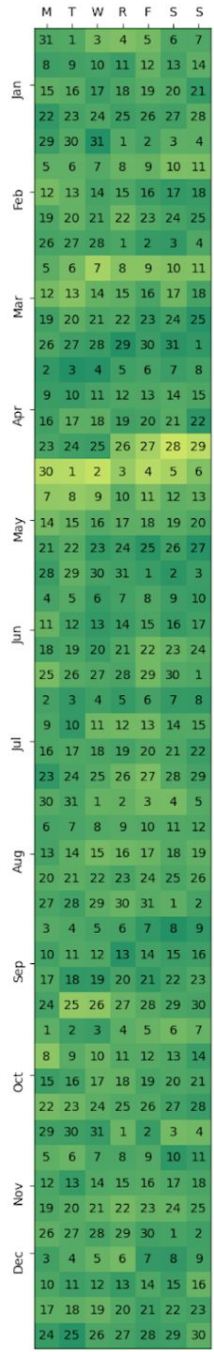
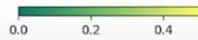
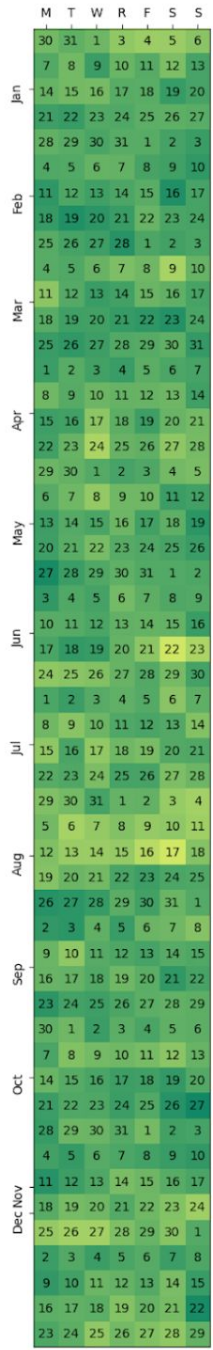
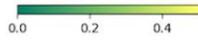

**Figure 4**. **Daily GBV probabilities.** GBV news heatmap for each day between 2018 to 2020. Each day is depicted with a color according to the probability of GBV news

**Subject analysis of gender-based violence news**

In the following, we focus on the intrinsic connection of subjects (or tags) in the news. We use the extracted GBV probability (outputted by the neural network NN2) to discern which news talks about this issue. We will consider a new to mainly cover gender violence if the probability returned by the neural network is greater than 0.9999, that is, if the neural network NN2 confirms it without doubt. This process leaves us with a set of 5375 news that cover gender violence in this way. We proceed then with a subject classification neural network (NN1) to tag this 5375 news and find out its main subjects.

Fig. 5 shows the resulting classification. The most common tags are 'justice', 'crime' and 'juries'. Such result can be explained because most of the news which cover gender violence or discrimination cases often also talk about laws or court cases.

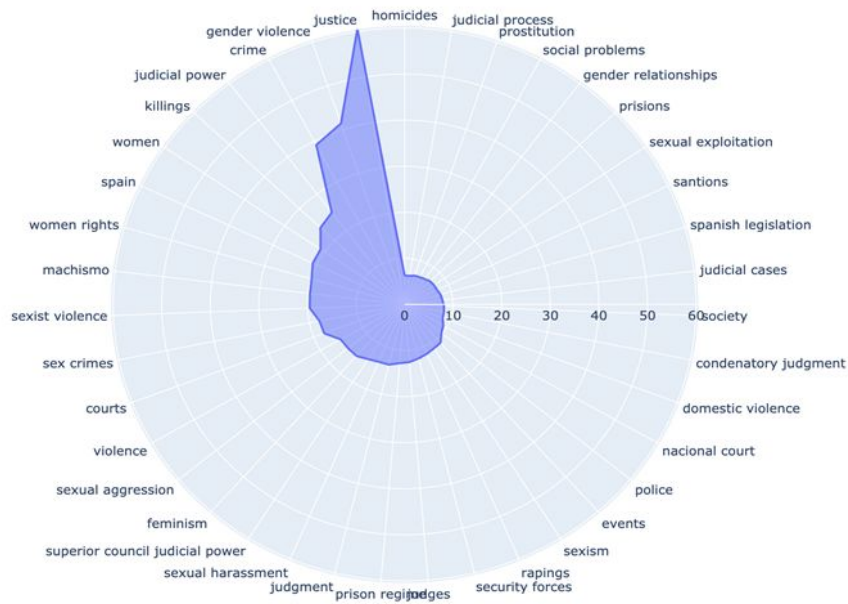

**Figure 5**. **Top topic classification of the gender-based violence news.** Subject frequency (percent) in GBV related news.

We use Topological Data Analysis[25] (TDA) (see methods for further details concerning the Mapper algorithm) to extract related topics. For that, we draw a graph that shows how news is connected in terms of their subjects. The diagram returned by Mapper algorithm was obtained following this process: First, we restrict the dataset selecting the news for a given year. For each news, we obtain the sequence of probabilities of each tag (using the subject classification neural network NN1), which gives us a sequence of vectors of more than 1000 dimensions (one for each possible tag). Second, we reduce the dimensionality of the previous sequence of vectors using principal component analysis. Then, we obtain a sequence of vectors (one for each new) with dimension 3 (the reduced dimension). This new sequence of vectors captures the geometric information of the thematic similarity of the news. Thirld, we apply the Mapper algorithm[26], which summarizes the closeness of the news with a geometric interpretation. The resulting graph draws a node for each cluster of news where the ones that are thematically close in terms of their subjects. Note that if these nodes are overlapped, i.e. there are news belonging to more than one cluster, the nodes will be joined with segments. This will allow us to see how subjects change and how news organize themselves in terms of closeness. Figure 6 shows the corresponding plot, where the warmer colors represent news (or clusters of them) with higher GBV probability.

In order to get insights about the figure, we check the branches of the graph. Each branch represents news' clusters whose subjects evolve similarly. As can be seen in Fig.6a, the GBV branch is the junction between feminism and politics. This findings fully accords with the central node of the GBV in regards to legislation and court cases. We repeat the analysis for 2020 news (see Fig.6b), where the GBV branch merges with the Justice one. Our findings are in accordance with the previous section results, where we show that justice subjects are highly related to GBV news.

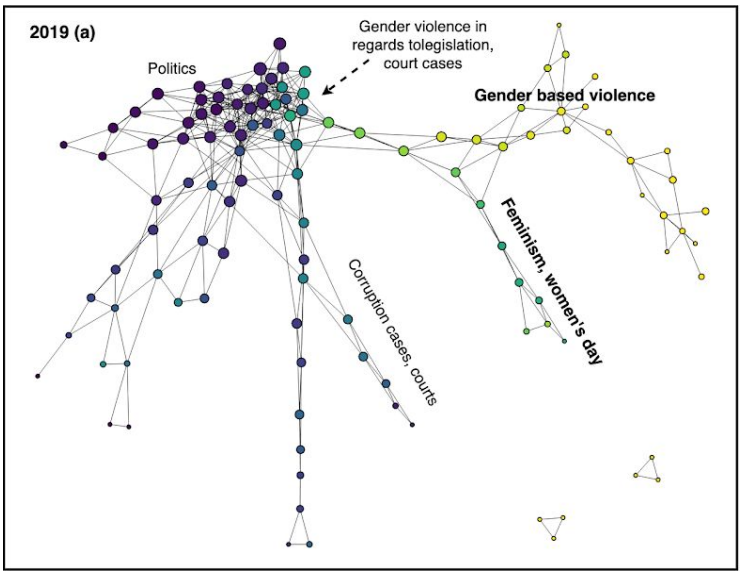
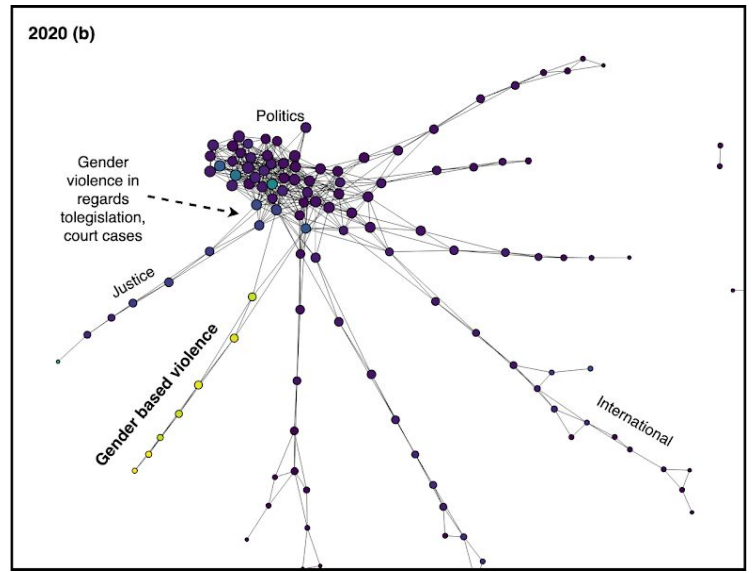

**Figure 6.Interdependence between topics.** Subject dependence. (left) 2019 and (right) 2020 diagram. Each node represents a cluster of news. Note that news belongs to the same cluster if they are closely related in terms of their subjectsI.e. the algorithm then connects clusters if they overlap. The colour of the nodes corresponds to the GBV probability (the higher GBV probability, the warmer colour is). The size of the nodes represents the amount of news in them.

**DISCUSSION**

Although our case study was limited to Spanish news, we suggest that the methodology proposed in this paper (rather than the specific results) might have a general validity. In fact, we show that Gender-Based Violence, the recent feminine movement and gender victims' voices have been in the spotlight lately in the news. The recent concern of GBV shown in society questionnaires, has resulted in a marked increase in publications about it. Specifically, this interdependence appears in the number of news about GBV and which echoes mediatic events. This result demonstrates an enhancement of the public perception of GBV

A strength of this Machine Learning study is the definition of a *GBV probability for each news*. We also tested how GBV propatility evolves over a certain period of time. For that, we map how GBV probability appears on a daily basis. After that, we pinpoint GBV related events that push both the increase in the news regarding this subject and public awareness. Finally, we used TDA techniques to extract how related certain topics (or tags) in the news. Our approach shows that GBV, usually unspoken as such, tends to be connected by the media with the subject "*justice*" and rarely with the subject "*feminism*". This finding may set a new point of view and representation of the news about GBV that would lead to better reporting. On this line, creation or improvement of already existing guidelines could be an interesting objective [27–29]. Hence, news media would help raise awareness on the problem that GBV poses, and thus, promote resources for its prevention. Furthermore, as the media reflects the concerns of our societies, it should help to neutralize GVB and combat it.

**METHODS**

- **DATASET EXTRACTION PROCESS**

We obtained nearly 784259 news from Spanish newspapers (from lavanguardia, elpais, elmundo, abc, 20minutos, publico and diario.es) from January 2005 to March 2020. To extract such an immense amount of news, we used a big data technique called web scraping (also called web crawling or mining). In particular, this technique gets a group of servers querying massive quantities of data from public web pages on the internet.

We deployed a network of cloud servers and a local server to break up the load of the work:

(i) A Mongodb database server: a no-sql database widely used in Big Data projects in order to store large amounts of data.

(ii) Two Cloud servers (scrapers): these network nodes contain our software coded with the Golang programming language. This software searches the web archive of the newspapers, then the nodes save the data in the Mongodb database.

(iii) A local server: A server with scripts coded in NodeJS and Python searches our Mongodb database and trains the neural networks to classify the subjects in the news. Then, this server analyzes the full database and by using neural network models calculates the probability that each new has of covering the subject of gender violence.

- **NATURAL LANGUAGE PROCESSING**

Natural Language Processing (NLP) techniques process unstructured (e.g. text) data to meaning data. The extracted corpus of documents has been cleaning and pre-processing: removing html tabs, removing accented characters, expanding contractions, removing special characters, lemmatization, removing stopwords, and tokenization. This process transformed news into structured vectorial data, which will be used in the developed neural network models.

We developed two convolutional neural network models to carry out the natural language processing of news:

> **(NN1)** We extract the topic of each news, which corresponds to a multilabel subject classification. For that we developed a convolutional neural network composed of an embedding layer, two convolutional (1 dimensional) layers, one pooling and one output dense layer. This neural network was trained with binary cross entropy loss of 0.0153 (multilabel accuracy 15%). Given a text input (news content for instance) this neural network outputs a set of tags (subjects) with their probabilities. Only tags with a probability higher than 0.5 are taken.

> **(NN2)** We extract how likely is that each news address GBV topics by a gender based violence binary classification. We developed a neural network composed of an embedding layer, four convolutional (1 dimensional) layers, two poolings and one output dense layer (with dimension equal to the number of possible tags). In this case we obtained binary cross entropy loss 0.0233 (accuracy of 98%, precision of 0.9884 and recall: 0.9865). This model associates each text input with a probability of GBV, that is, a 0-1 range where zero means lackness of GBV.

**Topological Data Analysis methods (Mapper algorithm)**

One key part of our analysis in the Topological data analysis (TDA) algorithm called Mapper. Since TDA is a relatively new branch of data analysis, we will introduce some of its principles and ideas. Topology is regarded as a pure or theoric branch of mathematics. The fundamental assumption in topology is that the way in which the parts of a system are connected is more important than distance. This particular approach is important to why topology is useful for analyzing and visualizing data.

Topological Data Analysis (TDA) is a modern discipline that uses the techniques of topology to analyze data. Mapper is a TDA algorithm that gives us a way to construct a graph (or simplicial complex) from data in a way that reveals some of the topological features of the space. It was developed by [25].

Though not exact, mapper is able to estimate important connectivity aspects of the underlying space of the data so that it can be explored in a visual format. We used the python implementation of the Mapper algorithm Kepler Mapper to study the connectivity of gender-based violence news and the rest of the news within a period [26].

**Data availability**
The data that support the findings of this study are available from the corresponding author upon reasonable request.

**Code availability**
The source code used in this study is available at https://github.com/news-scrapers/news-scraper-subject-classifiers-model and https://github.com/domestic-violence-ai-studies/domestic-violence-news-classifier-spanish.

**Acknowledgments**

**Contributions**
Conceiving the study and the design: H.J.B., N.P., C.L. Annotations, creating the datasets, developing and training the models: H.J.B. Conducting the statistical analysis and data analysis: H.J.B., C.L. Providing validation datasets: H.J.B. Data interpretation: H.J.B., N.P., C.L. Drafting the manuscript: H.J.B., C.L. Critically revising for important intellectual content: H.J.B., N.P., E.G., L.J.N, C.L.

**Corresponding author**

Correspondence to Hugo J. Bello.

**Competing interests**

The authors declare no competing interests.

8. Menéndez, M. I. & Menéndez, M. I. M. Retos periodísticos ante la violencia de género. El caso de la prensa local en España. *Comunicación y Sociedad* 53–77 (2014) doi:10.32870/cys.v0i22.48.

9. Fernández Teruelo, J. G. Feminicidios de género: Evolución real del fenómeno, el suicidio del agresor y la incidencia del tratamiento mediático. *Revista Española de Investigación* (2011).

10. Contributors to Wikimedia projects. La Manada rape case. https://en.wikipedia.org/wiki/La_Manada_rape_case (2018).

11. London School of Economics & Political Science. Artificial intelligence could help protect victims of domestic violence. https://www.lse.ac.uk/News/Latest-news-from-LSE/2020/b-Feb-20/Artificial-intelligence-could-help-protect-victims-of-domestic-violence.aspx.

12. Chapman, W. W. *et al.* Overcoming barriers to NLP for clinical text: the role of shared tasks and the need for additional creative solutions. *Journal of the American Medical Informatics Association: JAMIA* vol. 18 540–543 (2011).

13. Coppersmith, G., Leary, R., Crutchley, P. & Fine, A. Natural Language Processing of Social Media as Screening for Suicide Risk. *Biomed. Inform. Insights* **10**, 1178222618792860 (2018).

14. Farzindar, A. & Inkpen, D. *Natural Language Processing for Social Media: Second Edition*. (Morgan & Claypool Publishers, 2017).

15. Mori, K. & Haruno, M. Differential ability of network and natural language information on social media to predict interpersonal and mental health traits. *J. Pers.* (2020)
15

doi:10.2312/SPBG/SPBG07/091-100.

26. van Veen, H., Saul, N., Eargle, D. & Mangham, S. Kepler Mapper: A flexible Python implementation of the Mapper algorithm. *Journal of Open Source Software* vol. 4 1315 (2019).

27. Media coverage of gender-based violence - Handbook and Training of Trainers. https://eca.unwomen.org/en/digital-library/publications/2017/09/media-coverage-of-gender-based-violence---handbook-and-training-of-trainers.

28. [No title]. https://unesdoc.unesco.org/ark:/48223/pf0000371524.

29. Macharia, S. & Morinière, P. Learning resource kit on gender ethical journalism and media house policy. *World Association for Christian Communication & International Federation of Journalists, Toronto and Brussels* (2012).17